\documentclass[letterpaper, 10 pt, conference]{ieeeconf}  

\IEEEoverridecommandlockouts                              

\usepackage[l2tabu,orthodox]{nag}

\usepackage[pdftex,colorlinks]{hyperref}

\usepackage[printonlyused]{acronym}

\usepackage{siunitx}
\usepackage[all]{nowidow}

\usepackage{tikz}
\usetikzlibrary{arrows}
\usetikzlibrary{arrows.meta}
\usetikzlibrary{calc}
\usepackage{makecell}

\usepackage{lipsum}
\usepackage{comment}
\usepackage{algpseudocode}

\usepackage{xspace} 

\usepackage{subcaption} 
\usepackage{epstopdf}

\usepackage{import}

\usepackage{xcolor}
\usepackage{colortbl}
\usepackage{adjustbox}
\usepackage{hhline}
\usepackage{booktabs}

\usepackage{tabularx}
\usepackage{multirow, multicol}

\usepackage{rotating}

\usepackage[ruled,vlined,linesnumbered]{algorithm2e}

\usepackage{amssymb,amsfonts,amsmath,amscd}

\usepackage{amsthm}
\usepackage{mathalpha}

\theoremstyle{remark}
\newtheorem{remark}{Remark}

\usepackage{bm}

\newcommand{\bbm}{\begin{bmatrix}}
\newcommand{\ebm}{\end{bmatrix}}

\newcommand{\ignore}[1]{}

\newcommand{\bma}[1]{\left[\begin{array}{#1}}
\newcommand{\ema}{\end{array}\right]}

\DeclareMathAlphabet{\mbf}{OT1}{ptm}{b}{n}

\def\fdotb{{\raisebox{-0.6ex}{ \kern0.2ex\raisebox{0.8ex}{\tiny $\hspace*{-1ex}\circ$}}}}
\def\fddotb{{\raisebox{-0.6ex}{ \kern0.2ex\raisebox{0.8ex}{\tiny $\hspace*{-1ex}\circ\circ$}}}}

\newcommand{\utimes}{ {\raisebox{-0.6ex}{ \kern-1.0ex\raisebox{0.6ex}{ \small $\mathsf{v}$}}} } %
\newcommand{\beq}{\begin{equation}}
\newcommand{\eeq}{\end{equation}}
\newcommand{\bdis}{\begin{displaymath}}
\newcommand{\edis}{\end{displaymath}}
\newcommand{\beqarray}{\begin{eqnarray}}
\newcommand{\eeqarray}{\end{eqnarray}}
\newcommand{\beqarraynn}{\begin{eqnarray*}}
\newcommand{\eeqarraynn}{\end{eqnarray*}}

\DeclareMathAlphabet{\mbf}{OT1}{ptm}{b}{n}

\usepackage{balance}
\usepackage[T1]{fontenc}

\newcommand{\MainAlgAcc}{Informed BLT*}
\newcommand{\MainAlg}{Informed Belief Localization Trees*}
\newcommand{\MainAlgAccUninformed}{BLT*}
\title{\LARGE \bf
BLT*: Informed Belief Localization Trees for Uncertainty-Aware Planning on Digital Twins
}

\author{Elliot Preston-Krebs$^{1}$, Abhishek Goudar$^{1}$, and Timothy D. Barfoot$^{1}$%
\thanks{$^{1}$University of Toronto Institute for Aerospace Studies (UTIAS),
4925 Dufferin St, Ontario, Canada.
{\tt \{elliot.prestonkrebs, abhishek.goudar\}@robotics.utias.utoronto.ca,}
{\tt tim.barfoot@utoronto.ca}}%
}

\begin{document}
\bstctlcite{IEEEexample:BSTcontrol}

\maketitle
\thispagestyle{empty}
\pagestyle{empty}
\begin{abstract}
We present \MainAlg\ (\MainAlgAcc), a sampling-based belief space planning (BSP) algorithm that scales to large outdoor digital twins with point-cloud observations. We adapt RRT* and Informed RRT* to belief space using the $2$-Wasserstein ($W_2$) metric. 
Assuming isotropic Gaussian beliefs, sampled belief states can be connected efficiently while accounting for available information and probabilistic collision constraints. This enables steering and rewiring without repeatedly propagating observations, and allows previously computed measurement information to be reused.
We present a framework to generate semantically labelled digital twins for planning in real-world environments with point-cloud-based localization. Experiments in simulated environments and digital twins show faster initial solution discovery in most maps with competitive cost convergence.
\end{abstract}


\section{Introduction}
Autonomous unmanned ground vehicles (UGVs) operating in large mapped environments must plan not only where they can travel, but also where they can reliably localize. A geometrically short path may traverse poorly observable regions, causing uncertainty to grow until probabilistic collision-avoidance or goal-reaching constraints are no longer satisfied. Belief space planning (BSP) addresses this coupling by planning over distributions of robot states and jointly accounting for motion and localization uncertainty.

Sampling-based BSP methods either propagate beliefs over nominal configuration-space graphs~\cite{RRBT,IBBT} or sample directly in belief space~\cite{beliefA}. Although the $2$-Wasserstein ($W_2$) distance provides a suitable metric for Gaussian beliefs~\cite{W2Dist}, existing sampling-based methods use it primarily to define proximity while retaining a separate trajectory-cost objective. Other perception-aware planners incorporate spatial localizability or photometric information~\cite{schirmer2017efficient,costante2016perception}; however, sampling-based BSP is still commonly evaluated in low-dimensional environments with analytic or region-based observation models. Applying these methods to digital twins is more computationally demanding because localization quality must be inferred from local point-cloud geometry, while dynamic objects, vegetation, and reconstruction artifacts may provide unreliable observations.
%
\begin{figure}[t]
    \centering
    \includegraphics[
        width=\columnwidth
    ]{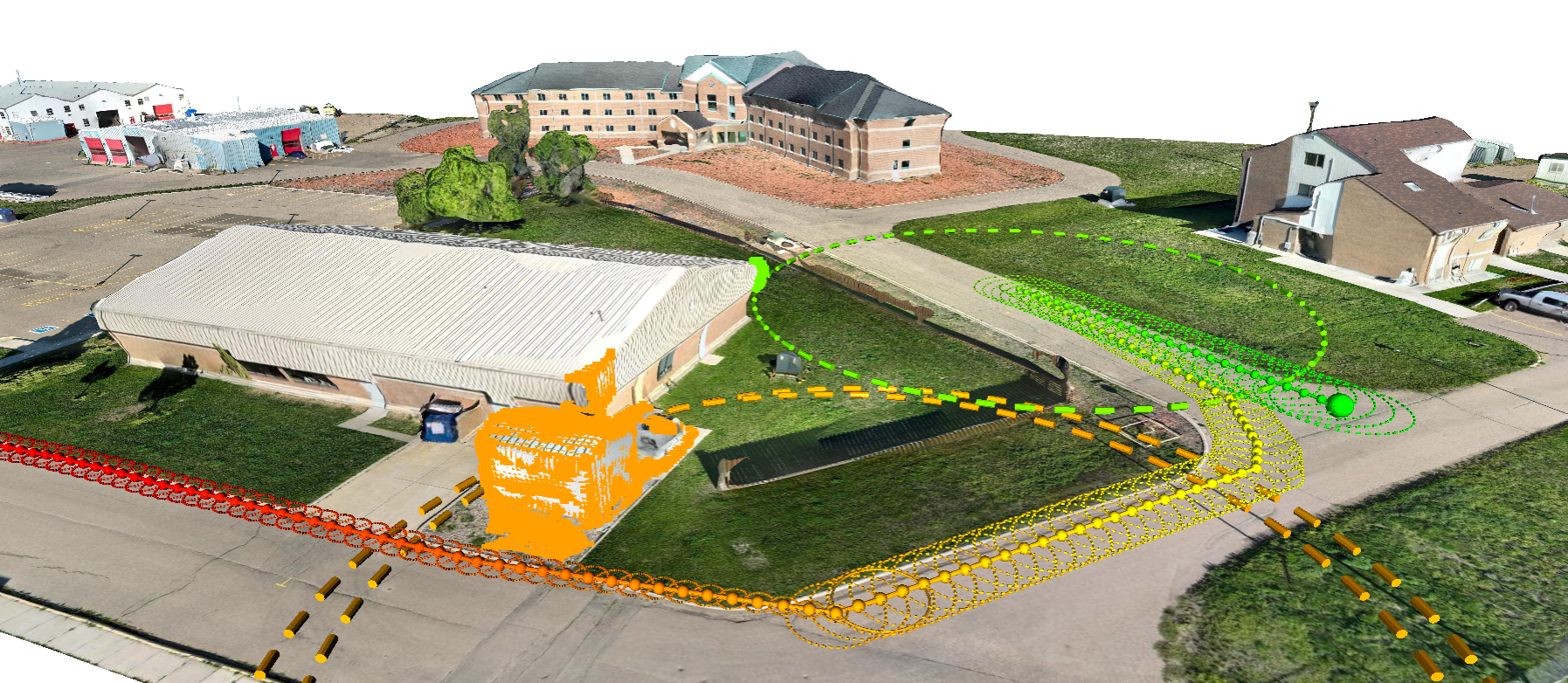}
    \includegraphics[
        width=\columnwidth,
        trim={1.5cm 7cm 1.5cm 1.5cm},
        clip
    ]{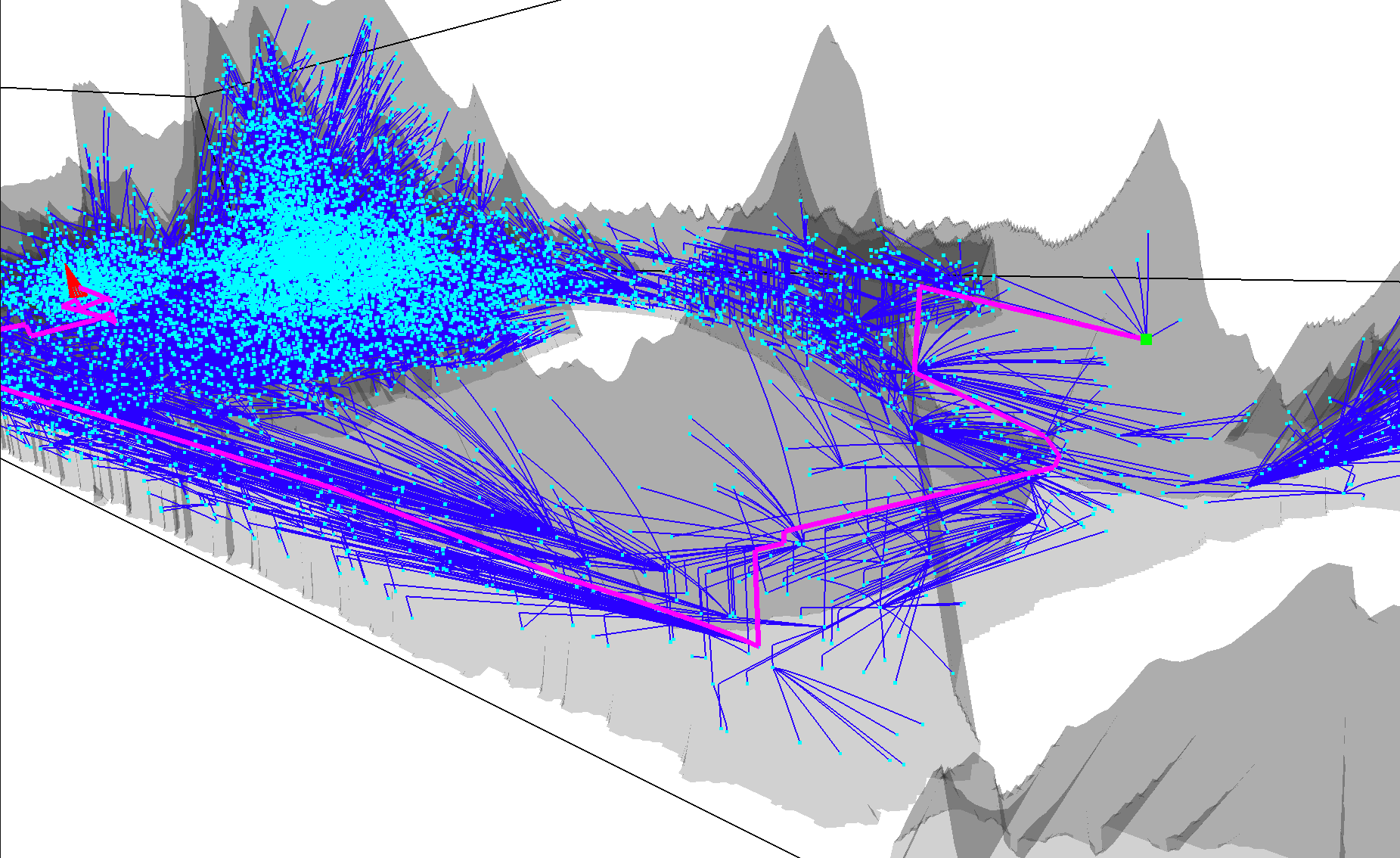}
    \caption{(Top) Configuration-space view of an Informed BLT* solution projected onto the \emph{Office} map. 
    The trajectory is coloured from green to red, indicating progression; point-cloud observations are colour-matched to their corresponding positions, and the ring denotes the lidar range. (Bottom) Belief-space planning view with uncertainty on the vertical axis, scaled for visualization. Grey surfaces are obstacles lifted into belief space under the uncertainty-based collision criterion in Section~\ref{sec:Interactions}. Blue edges and cyan nodes show the planning tree and the final path in magenta. The robot first detours within lidar range of a building to reduce uncertainty, allowing passage through a narrower corridor to a shorter path.}
    \label{fig:main}
    \vspace{-3mm}
\end{figure}
We present \MainAlg\ (\MainAlgAcc), which builds on RRT*~\cite{RRTstar} and Informed RRT*~\cite{gammell2014informed} to minimize accumulated $W_2$ path length in an isotropic Gaussian belief space. Under a holonomic motion model, we derive a closed-form belief-reachability condition that supports direct belief-space steering and rewiring without sampling and propagating control sequences for each state. Point-cloud information evaluated at existing nodes can also be reused during subsequent connection and rewiring.
Our contributions are: (i) belief-space reachability, steering, rewiring, and informed sampling under a $W_2$ path objective; (ii) a single prolate hyperspheroid outer approximation of the informed region induced by a convex goal set; and (iii) a digital-twin planning pipeline that uses semantic preprocessing to select localization-relevant geometry and the iterative closest point (ICP) Hessian as an information matrix for belief updates. We evaluate informed and uninformed variants of BLT* against three BSP baselines across 16 simulated and photogrammetry-reconstructed environments, comprising 8000 planner runs.
\section{Related Work}
Planning under uncertainty is commonly formulated as a partially observable Markov decision process (POMDP)~\cite{Littman_POMDP}, although exact solutions are generally intractable in continuous domains. BSP methods therefore employ various approximations. Censi et al.~\cite{censi2008bayesian} discretize the configuration space, propagate uncertainty using Fisher information, and retain non-dominated beliefs at each configuration. Platt et al.~\cite{platt2010belief} assume maximum-likelihood observations to obtain deterministic nominal belief trajectories amenable to linear-quadratic regulation, whereas Van den Berg et al.~\cite{van2012motion} use iterative linear-quadratic-Gaussian to optimize local trajectories and feedback policies without this assumption.

Graph-based approaches instead propagate uncertainty over sampled configuration-space structures. BRM~\cite{prentice2009belief} factors covariance propagation over a probabilistic roadmap~\cite{kavraki1996probabilistic}, while FIRM~\cite{agha2011firm} associates graph edges with feedback controllers to recover edge independence and the optimal-substructure properties for a graph search. CS-BRM~\cite{zheng2024cs} connects sampled belief nodes into a roadmap using covariance steering with finite-time reachability guarantees. RRBT~\cite{RRBT} bridges graph- and tree-based planning by incrementally constructing a graph of nominal trajectories and propagating non-dominated beliefs over its vertices; IBBT~\cite{IBBT} extends this approach with batch sampling and an admissible cost-to-go heuristic.

More recent methods sample directly in belief space. Belief-$\mathcal{A}$~\cite{beliefA} extends planners such as RRT~\cite{RRT} and SST~\cite{SST} to belief space and has been extended to continuous-time belief propagation and safety verification~\cite{mazouz2026continuous}. Littlefield et al.~\cite{W2Dist} establish continuity properties that support the use of the $W_2$ metric in optimal BSP. Whereas Belief-$\mathcal{A}$ uses $W_2$ for belief state proximity while retaining a separate trajectory cost, we use it as both the belief-space metric and path objective, enabling cost-informed sampling.

Environment-dependent localization has also been incorporated into planning via spatial localizability maps~\cite{schirmer2017efficient} and predicted photometric localization uncertainty~\cite{costante2016perception}. Our work instead extracts localization information from map geometry and incorporates it directly into the belief dynamics and $W_2$ objective for sampling-based planning in digital twins.
\section{Preliminaries} \label{sec:Preliminaries}
We use bold lowercase letters to denote vectors, e.g., $\mathbf{x} \in \mathbb{R}^d$, and bold uppercase letters for matrices, e.g., $\mathbf{A} \in \mathbb{R}^{d \times d}$. The space of $d \times d$ positive semidefinite matrices is denoted $\mathbb{S}^d_+$.

We denote the configuration space by $X \subseteq \mathbb{R}^d$; the obstacle space by $X_{\text{obs}} \subset X$; and the free space by $X_{\text{free}} = \operatorname{cl}(X \setminus X_{\text{obs}})$, where $\operatorname{cl}(\cdot)$ denotes closure.
Let $\mathbf{x}_{\text{start}} \in X_{\text{free}}$ denote the start state, and let $X_{\text{goal}} \subseteq X_{\text{free}}$ denote a closed, convex goal region.
We represent belief states by Gaussian distributions $b_i = \mathcal{N}(\mathbf{x}_i, \mathbf{\Sigma}_i)$, parameterized by $(\mathbf{x}_i, \mathbf{\Sigma}_i) \in \mathcal{B}^d$, where $\mathcal{B}^d \triangleq \mathbb{R}^d \times \mathbb{S}^d_+$ denotes the belief space.
Analogous to the Euclidean metric on $X$, we adopt the $W_2$ distance as the metric on $\mathcal{B}^d$ \cite{W2Dist}, where the squared $W_2$ distance between two Gaussian belief states is:
\begin{multline}
    W_2(b_1, b_2)^2 = \| \mathbf{x}_1 - \mathbf{x}_2\|_2^2 \\ 
    + \operatorname{tr}\left( \mathbf{\Sigma}_1 + \mathbf{\Sigma}_2 - 2 \left[ \mathbf{\Sigma}_1^{\frac{1}{2}} \mathbf{\Sigma}_2 \mathbf{\Sigma}_1^{\frac{1}{2}} \right]^{\frac{1}{2}} \right),
    \label{eq:W2}
\end{multline}
where $\|\cdot\|_2$ denotes the $\ell_2$ norm and $\operatorname{tr}(\cdot)$ denotes the trace operator.
We denote a belief-space planning tree by $\mathcal{T}=(V,E)$, where $V \subset \mathcal{B}^d$ is the set of belief-state vertices and $E \subseteq V \times V$ is the set of directed edges connecting them.
\section{Problem Formulation}\label{sec:ProblemFormulation}
\subsection{Belief Space Shortest Path Problem}
Consider the state space $X$, the belief space $\mathcal{B}^d$, the obstacle space $X_{\mathrm{obs}}$, and the goal region $X_{\mathrm{goal}}$ as defined in Section~\ref{sec:Preliminaries}. Let $P_{\mathrm{inter}}\in(0,1)$ denote a prescribed probability threshold for goal attainment and collision avoidance. Let $\beta:[0,1]\rightarrow\mathcal{B}^d$ denote a continuous belief-space path, where $\beta(t)=\mathcal{N}(\mathbf{x}_t,\mathbf{\Sigma}_t)$, and let $\Pi_{\mathrm{feas}}$ denote the set of paths consistent with the robot's belief dynamics, induced by the motion and observation models. Our objective is to find a feasible path from $b_{\mathrm{start}} =\mathcal{N}(\mathbf{x}_{\mathrm{start}}, \mathbf{\Sigma}_{\mathrm{start}})$ that minimizes its total length under the $W_2$ metric while satisfying the goal and collision-avoidance constraints:
\begin{equation} 
\begin{aligned}
    \beta^* ={}& \arg\min_{\beta\in\Pi_{\mathrm{feas}}} s(\beta) \\ \text{s.t.}\quad & \beta(0)=b_{\mathrm{start}}, \\ & \Pr_{\mathbf{x}\sim\beta(1)} \!\left(\mathbf{x}\in X_{\mathrm{goal}}\right) \geq P_{\mathrm{inter}}, \\ & \Pr_{\mathbf{x}\sim\beta(t)} \!\left(\mathbf{x}\in X_{\mathrm{free}}\right) \geq P_{\mathrm{inter}}, \qquad \forall t\in[0,1],
\end{aligned}
\end{equation} %
where the total $W_2$ path length is 
\begin{equation} 
    s(\beta) \triangleq \sup_{\substack{K\in\mathbb{N}\\ 0=t_0<t_1<\cdots<t_K=1}} \sum_{k=1}^{K} W_2\!\left(\beta(t_{k-1}),\beta(t_k)\right). 
\end{equation}
\subsection{System Dynamics}
Consider discrete-time motion and measurement models of the form
\begin{equation}
    \begin{aligned}
    \mathbf{x}_k &= \mathbf{f}(\mathbf{x}_{k-1}, \mathbf{u}_k) + \mathbf{w}_k, \quad &\mathbf{w}_k \sim \mathcal{N}(\mathbf{0}, \mathbf{Q}_k), \\
    \mathbf{y}_k &= \mathbf{g}(\mathbf{x}_{k}) + \mathbf{n}_k, \quad &\mathbf{n}_k \sim \mathcal{N}(\mathbf{0}, \mathbf{R}_k),
    \end{aligned}
    \label{eq:nonlinear_system}
\end{equation}
where $\mathbf{w}_k$ and $\mathbf{n}_k$ are the process and observation noises with covariances $\mathbf{Q}_k$ and $\mathbf{R}_k$, respectively, and $\mathbf{u}_k$ is the control input to the process model. For a holonomic system in $\mathbb{R}^2$, the model~\eqref{eq:nonlinear_system} can be expressed as the following linear motion and observation models: 
\begin{align}
    \mathbf{x}_k &= \mathbf{x}_{k-1} + \Delta\tau \mathbf{u}_k + \mathbf{w}_k, \label{eq:LinearMotionModel}\\
    \mathbf{y}_k &= \mathbf{x}_{k} + \mathbf{n}_k, \label{eq:LinearObservationModel}
\end{align}
where $\Delta\tau$ is a constant time step and $\mathbf{u}_k$ is the velocity.
\subsection{Isotropic Gaussian Assumption} \label{sec:isotropicAssumption}
We restrict the planner to isotropic Gaussian beliefs. When a prediction or observation update produces an anisotropic covariance, we conservatively bound it by an isotropic covariance determined by its maximum eigenvalue. Specifically, for a covariance matrix $\mathbf{\Sigma}$, we use the isotropic approximation
\begin{equation}
    \mathbf{\Sigma}_{\mathrm{iso}}
    =
    \lambda_{\max}(\mathbf{\Sigma})\mathbf{I},
\end{equation}
which satisfies $\mathbf{\Sigma}_{\mathrm{iso}} \succeq \mathbf{\Sigma}$. The corresponding belief space is
\begin{equation*}
    \mathcal{B}_{\mathrm{iso}}^d \triangleq \left\{ \mathcal{N}(\mathbf{x},\sigma^2\mathbf{I})  \mid \mathbf{x}\in\mathbb{R}^d,\, \sigma\geq0 \right\} \subset \mathcal{B}^d.
\end{equation*}
The benefit of this assumption is that the covariance of each state is parameterized by a single scalar standard deviation, $\mathbf{\Sigma}_i=\sigma_i^2\mathbf{I}$. We further assume isotropic process and observation noise, $\mathbf{Q}_k=\sigma_{q}^2\mathbf{I}$ and $\mathbf{R}_k=\sigma_{r}^2\mathbf{I}$, such that the state covariance remains isotropic under the system model.
The squared $W_2$ distance between two $d$-dimensional isotropic Gaussian beliefs is
\begin{align*}
    W_2(b_1, b_2)^2 &= \| \mathbf{x}_1 - \mathbf{x}_2\|_2^2 + d( \sigma_1^2 + \sigma_2^2 - 2\sigma_1\sigma_2) \\ &= \left\| \begin{bmatrix} \mathbf{x}_1^T & \sqrt{d}\sigma_1 \end{bmatrix}^{T} - \begin{bmatrix} \mathbf{x}_2^T & \sqrt{d}\sigma_2 \end{bmatrix}^{T} \right\|_2^2 \\ &= \ell_2 \left( \begin{bmatrix} \mathbf{x}_1^T & \sqrt{d}\sigma_1 \end{bmatrix}^{T}, \begin{bmatrix} \mathbf{x}_2^T & \sqrt{d}\sigma_2 \end{bmatrix}^{T} \right)^2.
\end{align*}
\begin{remark}[Belief Space Isometry] \label{remark:Isometry} 
The isotropic Gaussian belief space $\mathcal{B}_{\mathrm{iso}}^d$, equipped with the $W_2$ metric, admits an isometric embedding into $\mathbb{R}^{d+1}$ via $(\mathbf{x},\sigma) \mapsto [\mathbf{x}^T,\sqrt{d}\,\sigma]^T$.
In this work, we consider $d=2$, yielding $[x,y,\sigma]^T \mapsto [x,y,\sqrt{2}\,\sigma]^T$.
\end{remark} 
The holonomic motion model and isotropic uncertainty assumptions are sufficient for the use cases considered in our experiments.

\section{Methodology}\label{sec:Methodology}
In this section, we outline the evolution of belief states under our motion and observation models.
\subsection{Belief Dynamics}
We use the Kalman filter notation from \cite{barfoot2026state}, where $\check{(\cdot)}$ and $\hat{(\cdot)}$ denote prior and posterior quantities, respectively. The prediction step under the motion model~\eqref{eq:LinearMotionModel} is
\begin{equation}
\begin{aligned}
    \check{\mathbf{x}}_{k} &= \hat{\mathbf{x}}_{k-1} + \Delta\tau \mathbf{u}_k, \qquad
    &
    \check{\mathbf{\Sigma}}_{k} &= \hat{\mathbf{\Sigma}}_{k-1} + \mathbf{Q}_k.
\end{aligned}
\label{eq:LinearKFPredCov}
\end{equation}
Assuming the maximum-likelihood observation, the innovation is zero and the correction step is expressed in information form as
\begin{equation}
\begin{aligned}
    \hat{\mathbf{x}}_k &= \check{\mathbf{x}}_k, \qquad
    &\hat{\mathbf{\Sigma}}_k = \left( \check{\mathbf{\Sigma}}_k^{-1} + \boldsymbol{\Omega}_k\right)^{-1},
\end{aligned}
\label{eq:infocorrect}
\end{equation}
where $\boldsymbol{\Omega}_k=\mathbf{R}_k^{-1}$ is the information matrix in~\eqref{eq:LinearObservationModel}.
This form allows for information from point-cloud observations to efficiently be incorporated into our planning framework.
We bound the motion and observation covariances with the corresponding maximum eigenvalues to enforce isotropic covariances.
This is the tightest isotropic covariance bound on an anisotropic covariance, while respecting the minimum achievable uncertainty governed by the motion model and available information.
\begin{align}
    \check{\sigma}_{k}^{2}
    &= \lambda_{\max}\!\left(
        \hat{\mathbf{\Sigma}}_{k-1} + \mathbf{Q}_k
    \right)
    = \hat{\sigma}_{k-1}^{2} + \sigma_{q}^{2},
    \label{eq:IsotropicMotion}\\
    \hat{\sigma}_{k}^{2}
    &= \lambda_{\max}\!\left[
        \left(
            \check{\mathbf{\Sigma}}_{k}^{-1}
            + \boldsymbol{\Omega}_k
        \right)^{-1}
    \right] \notag\\
    &= \left(
        \check{\sigma}_{k}^{-2}
        + \lambda_{\min}(\boldsymbol{\Omega}_k)
    \right)^{-1},
    \label{eq:IsotropicCorrection}
\end{align}
where $\check{\mathbf{\Sigma}}_k = \check{\sigma}_k^2\mathbf{I}$, and $\hat{\mathbf{\Sigma}}_k = \hat{\sigma}_k^2\mathbf{I}.$
\subsection{Point-Cloud Observation Model} \label{sec:icp_observation}
In this work, we consider point-cloud observations in the correction step. We use the point-to-plane ICP \cite{pomerleau2013comparing} observation model. Under the maximum-likelihood observation assumption~\cite{platt2010belief}, the scan is aligned with the map at the predicted state, resulting in zero nominal residual. The ICP Hessian nevertheless provides a local approximation of the information associated with the observation.
Let $\mathbf{p}_j\in\mathbb{R}^3$ be a nominally aligned source point and $\mathbf{n}_j\in\mathbb{R}^3$ its normal, both expressed in the sensor frame at time step $k$. The stacked point-to-plane Jacobian and information matrix are
\begin{align}
    \mathbf{J}_k
    &=
    \begin{bmatrix}
        \mathbf{n}_1^{T} &
        (\mathbf{p}_1\times\mathbf{n}_1)^{T}\\
        \vdots & \vdots\\
        \mathbf{n}_M^{T} &
        (\mathbf{p}_M\times\mathbf{n}_M)^{T}
    \end{bmatrix},\\
    \mathbf{H}_k
    &=
    \mathbf{J}_k^{T}
    \mathbf{\Sigma}_r^{-1}
    \mathbf{J}_k,
    \qquad
    \mathbf{\Sigma}_r
    =
    \sigma_{r}^{2}\mathbf{I}_M.
\end{align}
Partitioning $\mathbf{H}_k$ such that $\mathbf{H}_{aa,k}\in\mathbb{R}^{2\times2}$ corresponds to the $xy$ translation components gives
\begin{equation}
    \boldsymbol{\Omega}_k
    =
    \mathbf{H}_{aa,k}
    -
    \mathbf{H}_{ab,k}
    \mathbf{H}_{bb,k}^{-1}
    \mathbf{H}_{ba,k},
    \label{eq:icp_marginal_information}
\end{equation}
where $\boldsymbol{\Omega}_k$ is the marginal planar information matrix used in~\eqref{eq:infocorrect}, assuming $\mathbf{H}_{bb,k}$ is nonsingular.

\subsection{Continuous Motion Model Parametrization}
From~\eqref{eq:IsotropicMotion}, variance accumulates linearly with discrete time steps, motivating a continuous-time interpolation between filter updates.
Assuming a fixed discrete update interval $\Delta\tau$, the variance at an arbitrary continuous time offset $\Delta t \geq 0$ from a prior belief state is given by
\begin{align*}
    \sigma_k^2 &= \sigma_{k-1}^2 + \sigma_{q}^2, &
    \sigma_2(\sigma_1, \Delta t) &= \sqrt{\sigma_1^2 +\frac{\Delta t \  \sigma_{q}^2}{\Delta\tau}}
\end{align*}
Since process noise accumulates uniformly at each discrete step independent of the control input, the total uncertainty incurred between two states is minimized by minimizing the number of prediction steps required to traverse them.
This is achieved by applying the maximum admissible control magnitude $u_{\max}$ directed along $\hat{\mathbf{d}}_{12} = \frac{\mathbf{x}_2 - \mathbf{x}_1}{\|\mathbf{x}_2 - \mathbf{x}_1\|_2}$, under a constant-velocity single-integrator motion model.
The state trajectory under this assumption is parameterized as
\begin{align*}
    \mathbf{x}_{k+1} &= \mathbf{x}_{k} + \Delta\tau\, u_{\max}\,\hat{\mathbf{d}}_{12}, \\
    \mathbf{x}_2(\Delta t) &= \mathbf{x}_1 + \Delta t\, u_{\max}\, \hat{\mathbf{d}}_{12}, \\
    \Delta t(\mathbf{x}_1, \mathbf{x}_2) &= \frac{\|\mathbf{x}_2 - \mathbf{x}_1\|_2}{u_{\max}}.
\end{align*}
Substituting $\Delta t(\mathbf{x}_1, \mathbf{x}_2)$ into $\sigma_2(\sigma_1, \Delta t)$ yields a closed-form lower bound on the reachable standard deviation at $\mathbf{x}_2$,
\begin{equation}
    \sigma_2(\sigma_1, \mathbf{x}_1, \mathbf{x}_2) = \sqrt{\sigma_1^2 + \kappa^* \|\mathbf{x}_2 - \mathbf{x}_1\|_2},
    \label{eq:reachable}
\end{equation}
where
\begin{equation}
    \kappa^* = \frac{\sigma_{q}^2}{\Delta\tau\, u_{\max}}
    \label{eq:Kstar}
\end{equation}
is the intrinsic inflation rate of the minimal motion model curve.
This bound is tight when the robot travels at $u_{\max}$ along $\hat{\mathbf{d}}_{12}$.

\subsection{Belief State Reachability \& Rewiring} \label{sec:reachability}
We gather candidate belief nodes for connection within a ball $\mathbb{B}(b_s,r_{\text{RRT}^*}) \subset \mathcal{B}_{\mathrm{iso}}^2$ using $\mathtt{Near}(\mathcal{T},b_s,r_{\text{RRT}^*})$. Following \cite{RRTstar}, the $W_2$ connection radius $r_{\text{RRT}^*}$ is determined from the number of nodes in the tree and the measure of the belief space, $\lambda(X)\sqrt{2}\,\sigma_{\max}$, where $\lambda(X)$ is the Lebesgue measure of the configuration space, used to approximate $\lambda(X_{\text{free}})$, and $\sigma_{\max}$ is the maximum sampled standard deviation. Remark~\ref{remark:Isometry} allows $W_2$ nearest-neighbour and volume calculations to use Euclidean geometry in the embedded belief space.

Our algorithm does not simulate control actions from existing beliefs. Instead, we sample $b_s$ directly from $\mathcal{B}_{\mathrm{iso}}^2$ and attempt to connect it to the existing tree.
A sample $b_s$ is reachable from a candidate parent $b_p$ without observations if $\sigma_s \geq \sigma_2(\sigma_p,\mathbf{x}_p,\mathbf{x}_s)$~\eqref{eq:reachable}.
When $\sigma_s$ exceeds this bound, we reparameterize the motion model curve with a higher inflation rate $\kappa > \kappa^*$,
\begin{equation}
    \kappa = \frac{\sigma_s^2 - \sigma_p^2}{\|\mathbf{x}_p - \mathbf{x}_s\|_2},
    \label{eq:inflation}
\end{equation}
which subsumes higher process noise, shorter time steps, or lower control velocity without resolving each individually.
When $\sigma_s$ is below this bound, reachability requires that the information $\boldsymbol{\Omega}_s$ at $\mathbf{x}_s$ can achieve an uncertainty lower than $\sigma_s$ via~\eqref{eq:IsotropicCorrection}.
In this case, we assume a partial measurement reduces uncertainty to exactly $\sigma_s$.
Both relaxations admit physically realizable interpretations: increased uncertainty may be obtained through zero-velocity propagation or increased process noise, while partial measurements may be approximated by a full update followed by stationary propagation to $\sigma_s$.
Collision validity is evaluated by sampling intermediate belief states along each motion edge at intervals $\Delta\tau$. For an edge using the reparameterized inflation rate in~\eqref{eq:inflation}, intermediate states are sampled using $\kappa$ rather than the minimal inflation rate $\kappa^*$ from~\eqref{eq:Kstar}.
The belief-state collision criterion is detailed in Section~\ref{sec:Interactions}. We denote the combined directed reachability and collision check by $\mathtt{ValidConnection}(b_p,b_s)$. A connection is valid if $b_s$ is reachable from $b_p$ under the belief dynamics above and the resulting belief trajectory is collision-free.

\subsection{Belief State Steering}\label{sec:steering}
Given a sample $b_s$, $\mathtt{Steer}(\mathcal{T},b_s)$ jointly selects a candidate parent $b_{\text{cand}}$ and generates a new belief $b_{\text{new}}$, returning $(b_{\text{cand}},b_{\text{new}})$.
Candidate parents are evaluated in two phases. First, the nearest node $b_{\text{nearest}}$ is evaluated. 
If it lies within the steering radius $r_{\text{steer}}$ and $b_s$ is reachable from $b_{\text{nearest}}$ according to Section~\ref{sec:reachability}, it is selected immediately.
Otherwise, all nodes within $\mathbb{B}(b_s,r_{\text{steer}})$ are searched, terminating when the first node from which $b_s$ is reachable is found.
Throughout the search, the candidate whose achievable post-observation uncertainty is closest to $\sigma_s$ is retained as a fallback.
If no candidate exactly reaches $b_s$, but a fallback exists, steering follows the minimal motion model curve and applies a full observation to generate the closest achievable belief $b_{\text{new}}$ to $b_s$.
If no candidate lies within the steering ball, $b_s$ is projected onto the boundary of the nearest node's steering ball before steering. The returned pair $(b_{\text{cand}},b_{\text{new}})$ is then accepted only if $\mathtt{ValidConnection}(b_{\text{cand}},b_{\text{new}})$; otherwise, the sample is rejected.

\subsection{Obstacle \& Goal Region Interaction} \label{sec:Interactions}
Consider a belief $b_s=\mathcal{N}(\mathbf{x}_s,\sigma_s^2\mathbf{I})$, and let $\mathbf{x} \sim b_s$.
We define
\begin{equation}
    k_{\text{inter}} = \sqrt{F_{\chi_d^2}^{-1} \!\left(P_{\text{inter}}\right)},
\end{equation}
where $F_{\chi_d^2}^{-1}$ is the inverse cumulative distribution function of a chi-squared random variable with $d$ degrees of freedom. The ball centred at $\mathbf{x}_s$ with radius $k_{\text{inter}}\,\sigma_s$ therefore contains $\mathbf{x}$ with probability $P_{\text{inter}}$.

Let $\mathtt{SDF}(\mathbf{x})$ denote the signed Euclidean clearance to $X_{\text{obs}}$, with positive values in free space. For a circular robot of radius $r_{\text{robot}}$, the sufficient collision-avoidance condition is
$\mathtt{SDF}(\mathbf{x}_s) \geq r_{\text{robot}} + k_{\text{inter}}\sigma_s$.
This condition guarantees that the robot is collision-free with probability at least $P_{\text{inter}}$. It is evaluated at the intermediate belief states generated by $\mathtt{ValidConnection}$, as described in Section~\ref{sec:reachability}.

Similarly, define the clearance of $\mathbf{x}_s$ from the exterior of the goal region as$d_{\text{goal}}(\mathbf{x}_s) = \operatorname{dist} \!\left(\mathbf{x}_s,X\setminus X_{\text{goal}}\right)$, where $\operatorname{dist}(\cdot)$ is the minimum distance between $\mathbf{x}_s$ and $X\setminus X_{\text{goal}}$.
The belief satisfies the goal condition with probability at least $P_{\text{inter}}$ if $d_{\text{goal}}(\mathbf{x}_s) \geq r_{\text{robot}} + k_{\text{inter}}\sigma_s$.
We denote this test by $\mathtt{InGoalRegion}(b_s)$.

\newcommand{\mapwithscale}[5][]{%
\makebox[#2][l]{%
\begin{tikzpicture}
    \node[anchor=south west, inner sep=0] (img) at (0,0)
        {\includegraphics[width=#2,trim={#1},clip]{#3}};

    \begin{scope}[x={(img.south east)}, y={(img.north west)}]
        \coordinate (a) at (0.07,0.07);
        \coordinate (b) at ($(a)+(#4,0)$);

        \draw[line width=1.2pt] (a) -- (b);
        \draw[line width=1.2pt] (a) ++(0,-0.012) -- ++(0,0.024);
        \draw[line width=1.2pt] (b) ++(0,-0.012) -- ++(0,0.024);
        \node[above] at ($(a)!0.5!(b)$) {\small #5};
    \end{scope}
\end{tikzpicture}%
}%
}
\begin{figure*}[t]
    \centering
    \vspace{-0.6em}
    \includegraphics[width=0.98\textwidth]{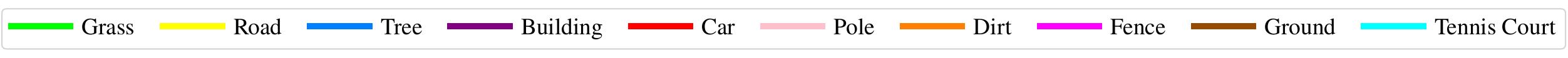}
    \vspace{-0.1em}
    
    \mapwithscale[0 0.39cm 0 0]{0.48\textwidth}{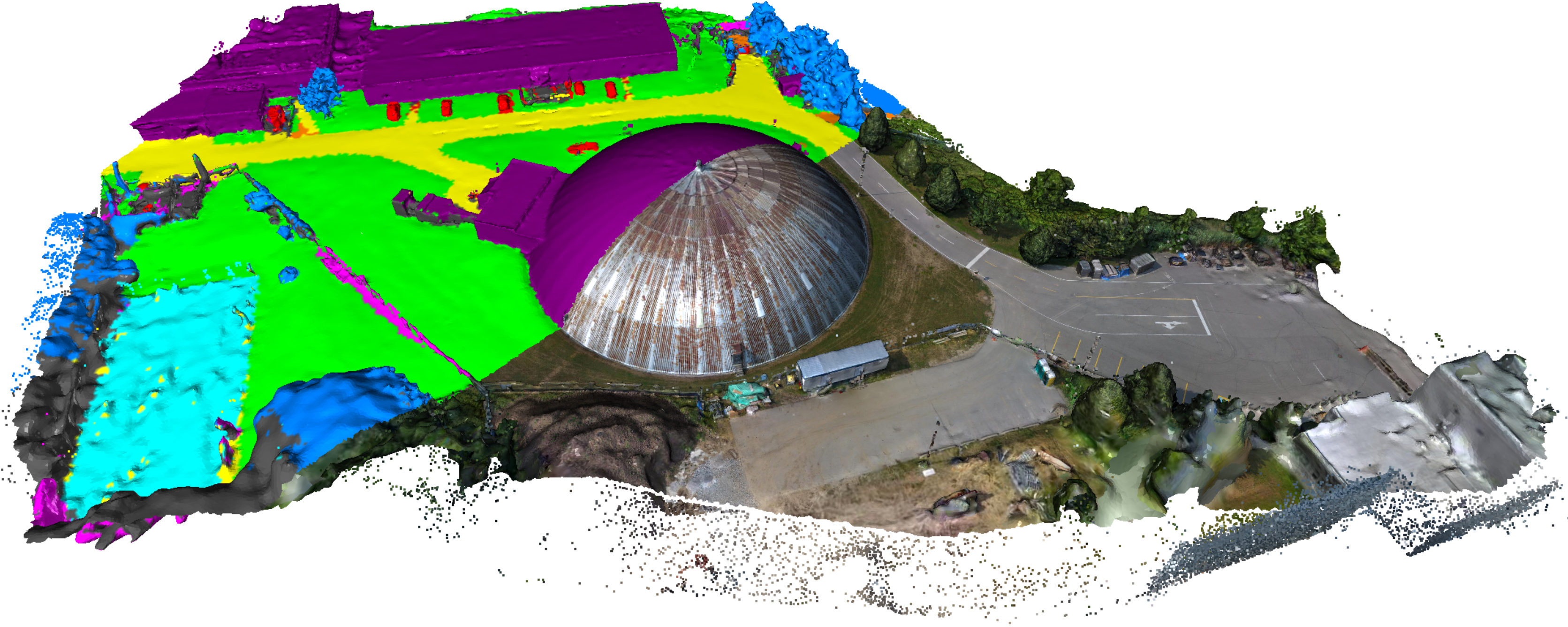}{0.2445}{50\,m}%
    \hfill
    \mapwithscale[0 0 0 0]{0.48\textwidth}{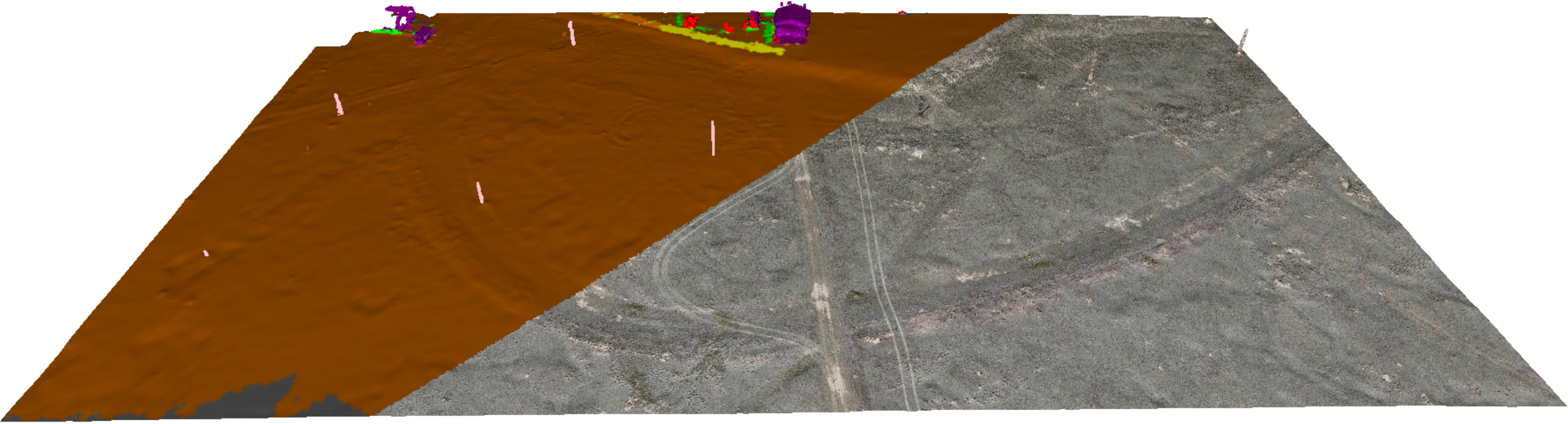}{0.1989}{50\,m}
    \mapwithscale[0 0 0 0]{\textwidth}{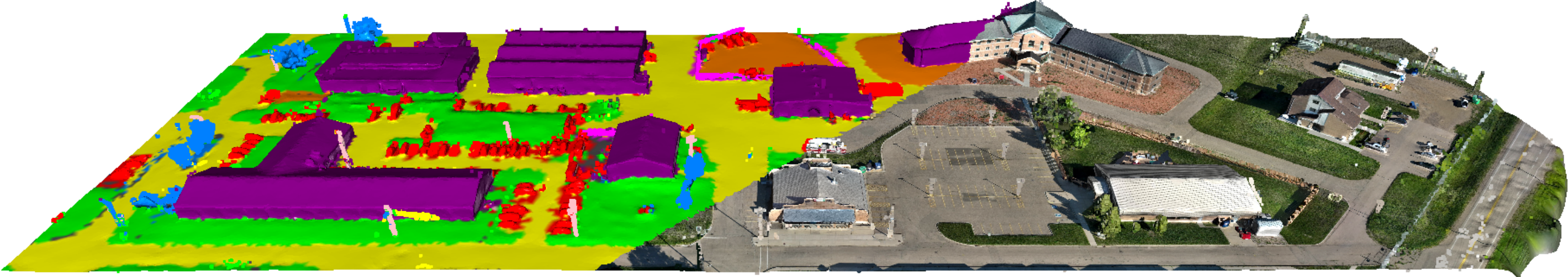}{0.2286}{100\,m}
    
    \vspace{-0.4em}
    \caption{Digital twins used for evaluation: Campus and Rural (top), and Office (bottom).}
    \label{fig:DTFIG}
    \vspace*{-1em}
\end{figure*}

\subsection{Informed Sampling for Convex Goal Region} \label{sec:informedSampling}
Let $X_{\text{goal}} \subseteq \mathbb{R}^d$ be a closed convex goal region, $\mathbf{x}_{\text{start}} \in \mathbb{R}^d$ the start state, and $c_i \in \mathbb{R}_{\ge0}$ the current solution cost (path length).
The true informed region is
\begin{equation}
\begin{aligned}
    X_{\text{informed}}
    = \bigl\{ \mathbf{x} \mid\;&
    \|\mathbf{x}_{\text{start}} - \mathbf{x}\|_2 \\
    &+ \min_{\mathbf{x}^\prime_{\text{goal}} \in X_{\text{goal}}}
    \|\mathbf{x} - \mathbf{x}^\prime_{\text{goal}}\|_2
    \le c_i \bigr\}.
\end{aligned}
\end{equation}
We denote $\mathtt{PHS}(\mathbf{x}_1,\mathbf{x}_2,c)$ as a function returning the prolate hyperspheroid (PHS) with foci $\mathbf{x}_1$ and $\mathbf{x}_2$ and major-axis length $c$. 
We seek a single PHS that approximates an outer bound on the informed region,
\begin{equation}
    X_{\text{informed,bound}} = \mathtt{PHS}(\mathbf{x}_{\text{start}}, \mathbf{x}^*_{\text{goal}}, c_{\text{informed}}).
\end{equation}

\subsubsection{Approximate informed region}
For any $\mathbf{y}\in\mathbb{R}^d$, define 
\begin{equation} 
    c(\mathbf{y}) = c_i + \max_{\mathbf{x}'_{\mathrm{goal}}\in X_{\mathrm{goal}}} \left\| \mathbf{x}'_{\mathrm{goal}}-\mathbf{y} \right\|_2. 
\end{equation} 
We use $\mathtt{PHS}(\mathbf{x}_{\mathrm{start}},\mathbf{y},c(\mathbf{y}))$ as an empirical outer approximation of $X_{\text{informed}}$. In our experiments, it provided sufficient samples in $X_{\text{informed}}$ to retain informed-sampling benefits. Establishing containment guarantees remains future work.

\subsubsection{Minimax tightening} We minimize $c(\mathbf{y})$, i.e., minimize the worst-case distance from $\mathbf{y}$ to $X_{\text{goal}}$:
\begin{equation}
\begin{aligned}
    \mathbf{x}_{\text{goal}}^*
    &= \arg\min_{\mathbf{y}}
    \max_{\mathbf{x}^\prime_{\text{goal}} \in X_{\text{goal}}}
    \|\mathbf{x}^\prime_{\text{goal}} - \mathbf{y}\|_2, \\
    c_{\text{interior}}
    &= \max_{\mathbf{x}^\prime_{\text{goal}} \in X_{\text{goal}}}
    \|\mathbf{x}^\prime_{\text{goal}} - \mathbf{x}_{\text{goal}}^*\|_2, \\
    c_{\text{informed}}
    &= c(\mathbf{x}_{\text{goal}}^*) =c_i + c_{\text{interior}}, \\
    X_{\text{informed,bound}} &= \mathtt{PHS}(\mathbf{x}_{\text{start}}, \mathbf{x}_{\text{goal}}^*, c_{\text{informed}}).
    \label{eq:informed_region_bound}
\end{aligned}
\end{equation}
$X_{\text{informed,bound}}$ minimizes major-axis length, with one focus at $\mathbf{x}_{\text{start}}$, but not necessarily volume. 

The configuration-space goal $X_{\mathrm{goal}}$, together with $P_{\mathrm{inter}}$, induces the corresponding belief-space goal region. Although derived in configuration space for clarity, Remark~\ref{remark:Isometry} allows the same PHS construction in the Euclidean embedding of $\mathcal{B}_{\mathrm{iso}}^2$. Accordingly, $\mathtt{Sample}(b_{\mathrm{start}},X_{\mathrm{goal}},c_i)$ in Algorithm~\ref{alg:main}, with $c_i$ the current solution cost, samples an $\mathbb{R}^3$ PHS in $[x,y,\sqrt{2}\sigma]$ coordinates following \cite{gammell2014informed,gammell2018informed}. 
Samples outside the true informed belief-space region, outside the belief-space bounds, or with $\sigma<0$ are rejected and resampled. Clamping or reflecting negative uncertainties would produce non-uniform samples.

\begin{algorithm}[t]
\small 
\caption{$\text{\MainAlgAcc}(b_{\mathrm{start}}, X_{\mathrm{goal}},X_{\mathrm{obs}})$} 
\label{alg:main}
$V \gets \{b_{\mathrm{start}}\}$; $E \gets \emptyset$; $\mathcal{B}_{\mathrm{soln}} \gets \emptyset$; $\mathcal{T} = (V,E)$\;
\For{$\mathrm{iteration}=1\ldots n$}{
    $c_{i} \gets \mathrm{min}_{b_{\mathrm{soln}} \in \mathcal{B}_{\mathrm{soln}}} \left\{\mathtt{Cost}(b_{\mathrm{soln}})\right\}$\;
    $b_{\mathrm{rand}} \gets \mathtt{Sample} (b_{\mathrm{start}}, X_{\mathrm{goal}}, c_{i})$\;
    $\textcolor{red}{(b_{\mathrm{cand}},b_{\mathrm{new}}) \gets \mathtt{Steer}(\mathcal{T},b_{\mathrm{rand}})}$\;
    \If{$\textcolor{red}{\mathtt{ValidConnection} (b_{\mathrm{cand}},b_{\mathrm{new}})}$}{
        $\mathcal{B}_{\mathrm{near}} \gets \mathtt{Near}(\mathcal{T},b_{\mathrm{new}},r_{\mathrm{RRT}^*})$\;
        $V \gets V \cup \{b_{\mathrm{new}}\}$\;
        $b_{\mathrm{min}} \gets b_{\mathrm{cand}}$\;
        $c_{\mathrm{min}} \gets \mathtt{Cost}(b_{\mathrm{min}}) + c(\mathtt{Connect}(b_{\mathrm{min}}, b_{\mathrm{new}}))$\;
        \For{$\forall b_{\mathrm{near}} \in \mathcal{B}_{\mathrm{near}}$}{
            $c_{\mathrm{new}} \gets \mathtt{Cost}(b_{\mathrm{near}}) + c ( \mathtt{Connect} (b_{\mathrm{near}},b_{\mathrm{new}}))$\;
            \lIf{$\textcolor{red}{\mathtt{ValidConnection}(b_{\mathrm{near}}, b_{\mathrm{new}})} \land c_{\mathrm{new}}<c_{\mathrm{min}}$}
            {$b_{\mathrm{min}} \gets b_{\mathrm{near}}$; $c_{\mathrm{min}} \gets c_{\mathrm{new}}$}
        }
        $E \gets E \cup \{(b_{\mathrm{min}},b_{\mathrm{new}})\}$\;
        \For{$\forall b_{\mathrm{near}} \in \mathcal{B}_{\mathrm{near}}$}{
            \If{$\textcolor{red}{\mathtt{ValidConnection}(b_{\mathrm{new}}, b_{\mathrm{near}})}$$\land \mathtt{Cost}(b_{\mathrm{new}}) + c ( \mathtt{Connect}(b_{\mathrm{new}},b_{\mathrm{near}})) < \mathtt{Cost}(b_{\mathrm{near}})$}{
                $b_{\mathrm{parent}} \gets \mathtt{Parent}(b_{\mathrm{near}})$\;
                $E \gets (E \setminus \{(b_{\mathrm{parent}},b_{\mathrm{near}})\}) \cup \{(b_{\mathrm{new}},b_{\mathrm{near}})\}$\;
            }
        }
        \If{$\mathtt{InGoalRegion}(b_{\mathrm{new}})$}{
            $\mathcal{B}_{\mathrm{soln}} \gets \mathcal{B}_{\mathrm{soln}} \cup \{b_{\mathrm{new}}\}$\;
        }
    }
}
\Return{$\mathcal{T}$}\;
\end{algorithm}

\subsection{\MainAlg}
We present \MainAlgAcc \ and its uninformed variant, \MainAlgAccUninformed. Both follow the general structure of RRT* \cite{RRTstar}, with steering, connection validity, and rewiring adapted to the belief dynamics described in Section~\ref{sec:reachability} and Section~\ref{sec:steering}. The informed variant additionally uses the sampling procedure of Section~\ref{sec:informedSampling} once a solution is available.

For a valid pair $(b_i,b_j)$, $\mathtt{Connect}(b_i,b_j)$ constructs the corresponding belief-space trajectory using the motion and observation models described above. Under the $W_2$ objective, the edge cost accumulates $W_2$ distance along the interpolated motion-model trajectory and across observation updates, since the associated reduction in uncertainty also induces motion through belief space. This preserves the full belief-space trajectory rather than collapsing motion and observation into a single edge.

Under the $\ell_2$ objective, edge cost is instead computed only from the configuration-space motion of the belief mean. Changes in uncertainty, including both its growth during motion and reduction during observations, do not contribute to the objective, although they continue to determine reachability, collision validity, and goal satisfaction.
$\mathtt{Cost}(b)$ denotes the accumulated edge cost from $b_{\text{start}}$ to $b$ in the tree. Parent selection and rewiring in Algorithm~\ref{alg:main} are therefore performed according to the selected $W_2$ or $\ell_2$ objective while sharing the same belief-space feasibility constraints.

\begin{figure*}[t]
    \centering

    \begin{subfigure}[t]{0.49\textwidth}
        \centering
        \includegraphics[width=\linewidth]{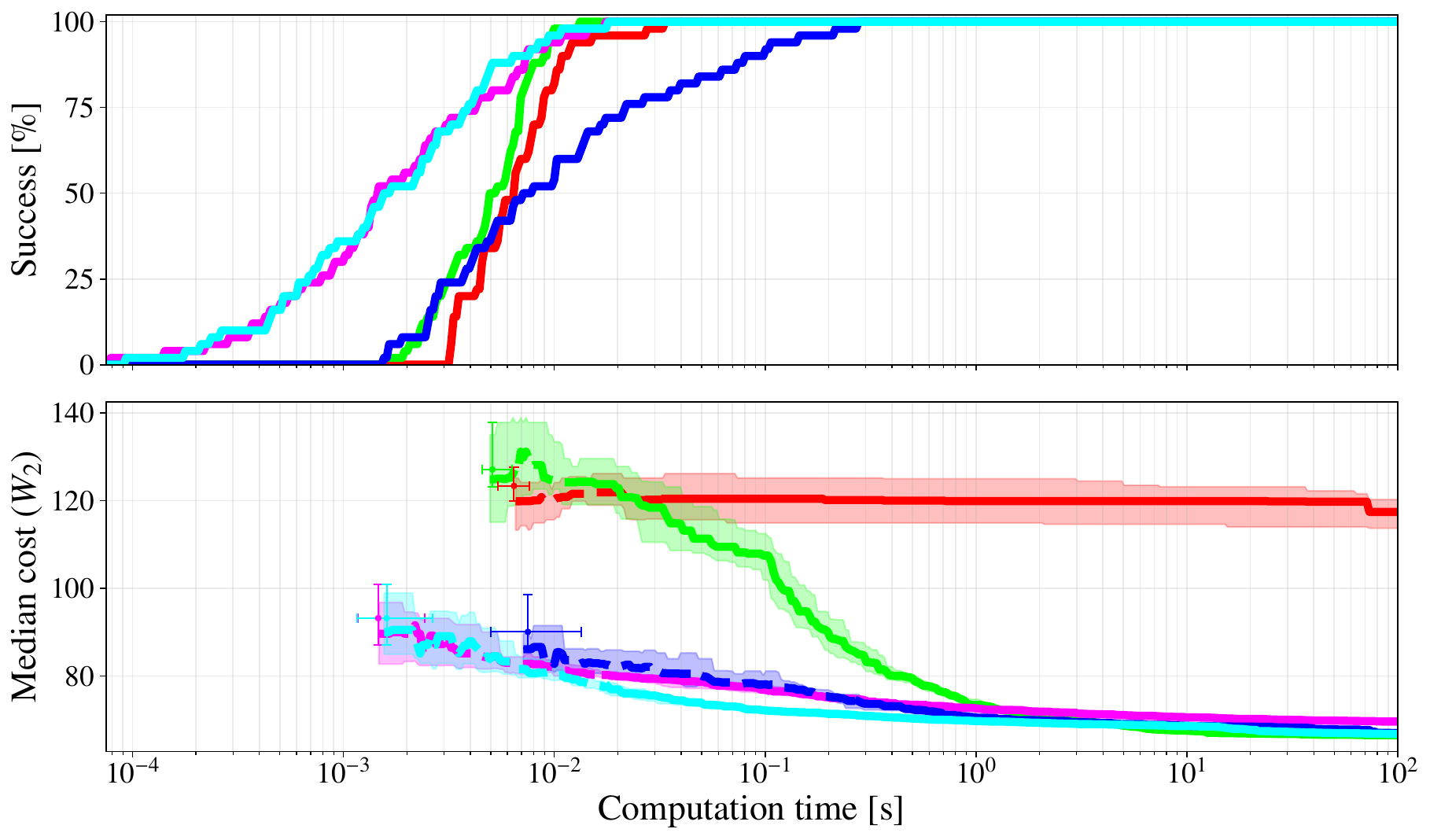}
        \caption{Random: Median cost and success versus time.}
        \label{fig:bench_block}
    \end{subfigure}
    \hfill
    \begin{subfigure}[t]{0.49\textwidth}
        \centering
        \includegraphics[width=\linewidth]{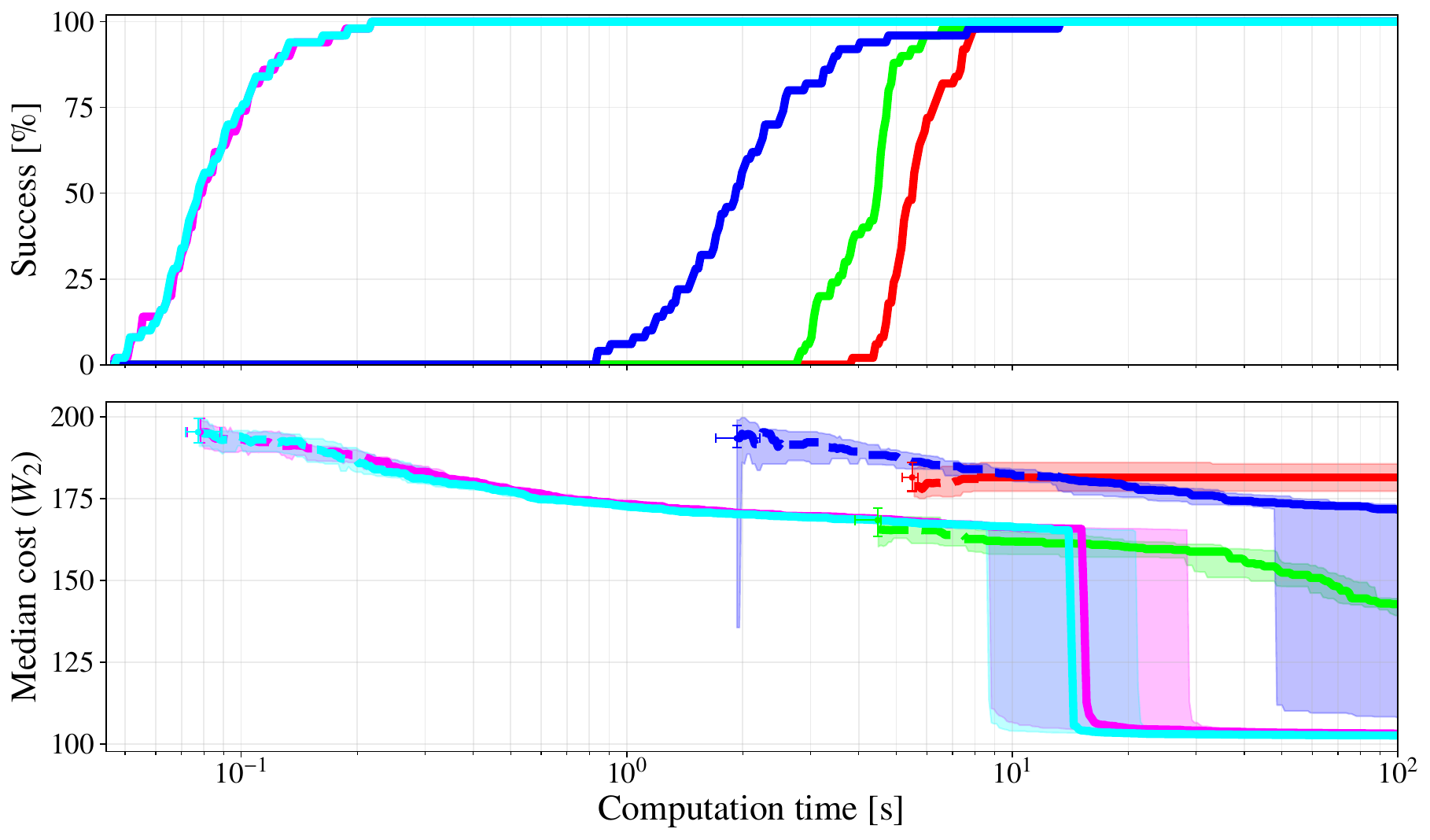}
        \caption{Campus: Median cost and success versus time.}
        \label{fig:bench_dome}
    \end{subfigure}

    \vspace{0.5em}
    \includegraphics[width=0.65\textwidth]{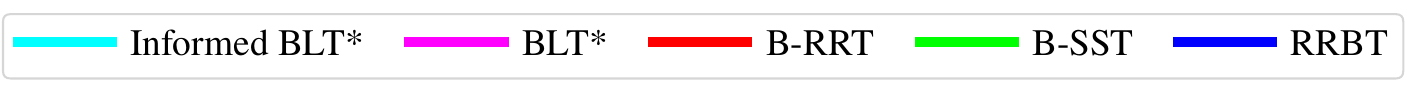}

    \caption{Benchmark results for \emph{Campus} and one representative \emph{Random} map over 50 independently seeded trials. Top: percentage of trials solved versus computation time. Bottom: median solution cost versus computation time, with non-parametric $95\%$ confidence bands. Dots denote the median initial solution, with $95\%$ confidence intervals on initial-solution time and cost.}
    \label{fig:bench}
    \vspace*{-1em}
\end{figure*}

\subsection{Digital Twin Preprocessing}
Our digital-twin preprocessing pipeline produces a 2.5D planning map $(X_{\text{free}},X_{\text{obs}})$ and extracts point clouds used for localization. It assigns user-defined semantic classes to the reconstructed geometry, which are designated as traversable, non-traversable, or observable for localization. Photogrammetric reconstruction quality varies across object types, so semantic class provides a proxy for localization reliability.

\subsubsection{Semantic Labelling}
We render the mesh together with point-cloud geometry not adequately represented by the mesh and segment each view using the open-vocabulary Segment Anything Model 3 (SAM3)~\cite{sam3}. 
The resulting masks are projected onto visible mesh faces and point-cloud points (see Fig.~\ref{fig:main}), and labels are assigned by majority vote across views.
Nearby points inherit the votes of their closest mesh faces, and unlabelled points may be infilled using distance-weighted neighbourhood voting. The user then selects the classes to be retained for ICP localization, typically favouring well-reconstructed static structures over transient or poorly reconstructed objects.

\subsubsection{2.5D Map Representations} \label{sec:MapRep}
We condense the planning space into a 2.5D map encoding height and occupancy, the latter as a Signed Distance Function (SDF). Occupancy ($X_{\text{free}}$, $X_{\text{obs}}$) is computed on a grid finer than the point cloud resolution ($5\,\mathrm{cm}$).
Cells containing a traversable-labelled point are marked free, cells containing only non-traversable points are marked obstacle, and cells with no points are marked unknown, with ties resolved in favour of obstacle. 

The height assigned to each free cell is the mean elevation of point-cloud points within it; nearby cells without observations inherit the height of the nearest observed cell. The cleaned occupancy grid is converted to $\mathtt{SDF}(\mathbf{x})$, positive in free space and negative inside obstacles, with unknown cells treated as obstacles. Both height and $\mathtt{SDF}(\mathbf{x})$ are queried continuously at any $\mathbf{x} \in X$ using bilinear interpolation.

\subsubsection{Integration with Planners}
During planning, the height map is queried at each candidate configuration to place the sensor at a fixed height above the terrain, and local point-cloud observations are formed from the semantic classes retained for ICP localization. The corresponding ICP information matrix $\boldsymbol{\Omega}$ is evaluated only after the motion to that configuration is found to be collision-free.
For both BLT* variants, the evaluated information matrix is stored with the corresponding tree node and reused during subsequent connection checks. The same caching is applied along edges in the RRBT graph. In contrast, B-RRT and B-SST generate new states through action sampling, which generally produces new configurations; reusing previously computed information would therefore require an additional spatial lookup and provide less direct benefit.

\begin{figure}[t]
    \vspace{0.6em}
    \centering
    \includegraphics[width=0.25\columnwidth]{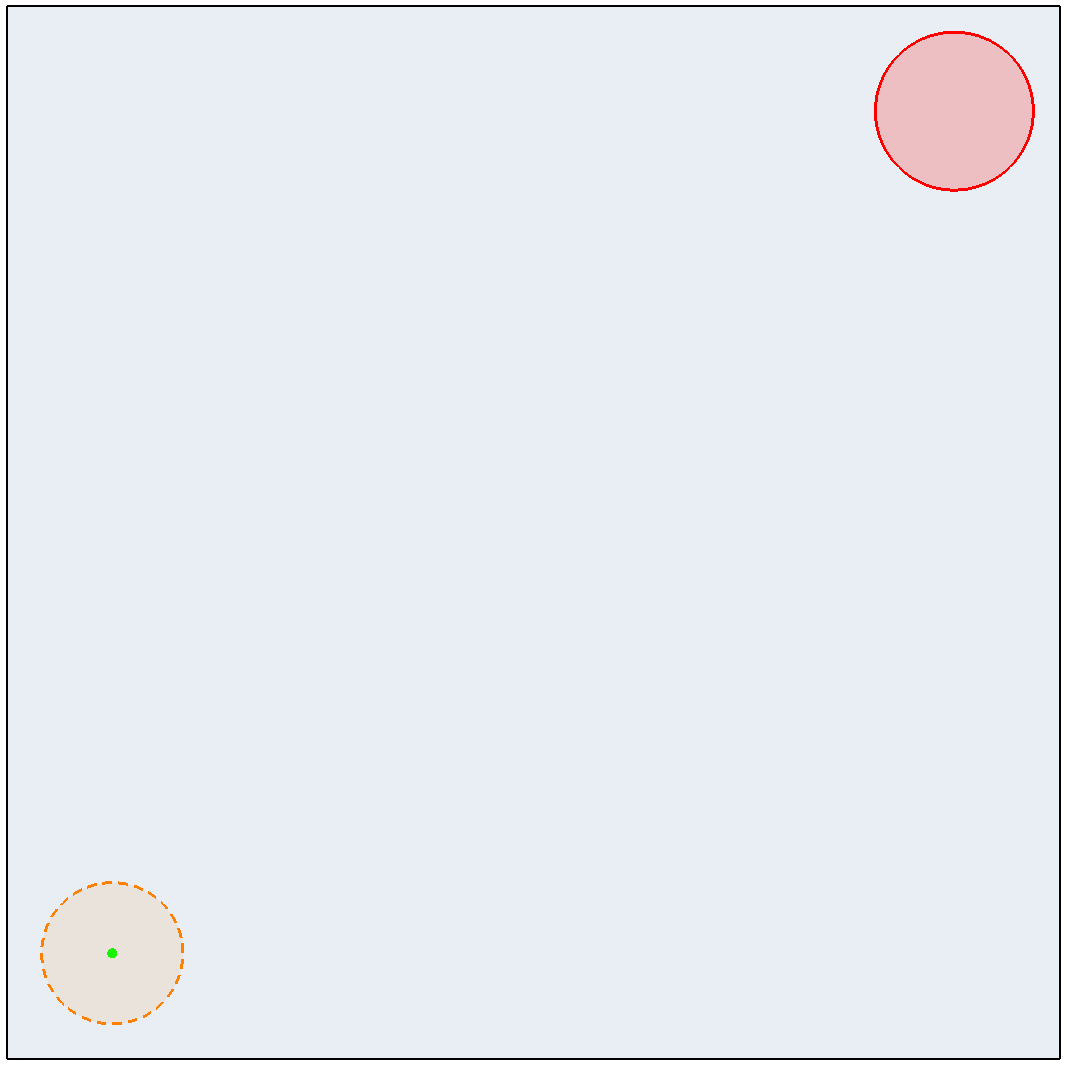}%
    \includegraphics[width=0.25\columnwidth]{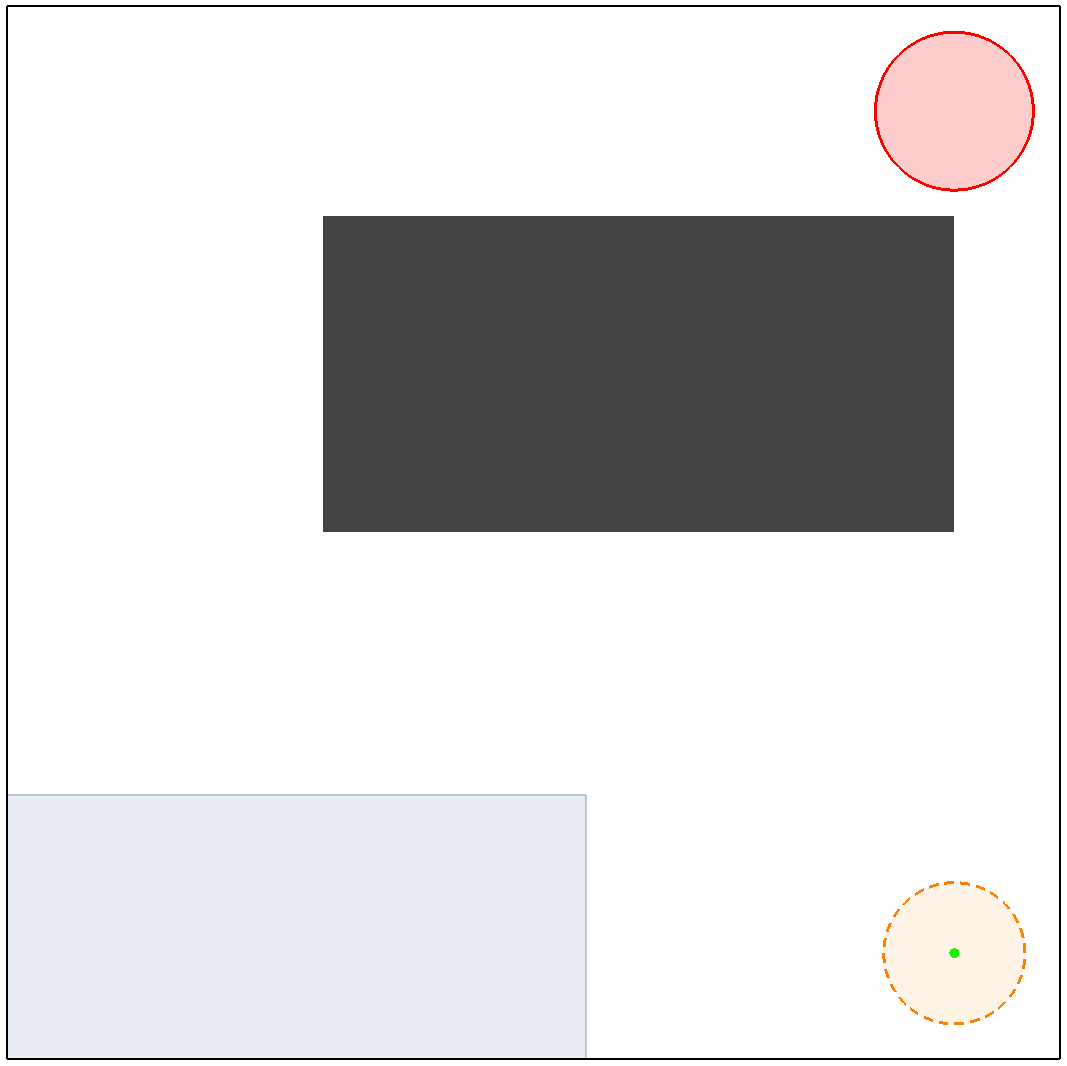}%
    \includegraphics[width=0.25\columnwidth]{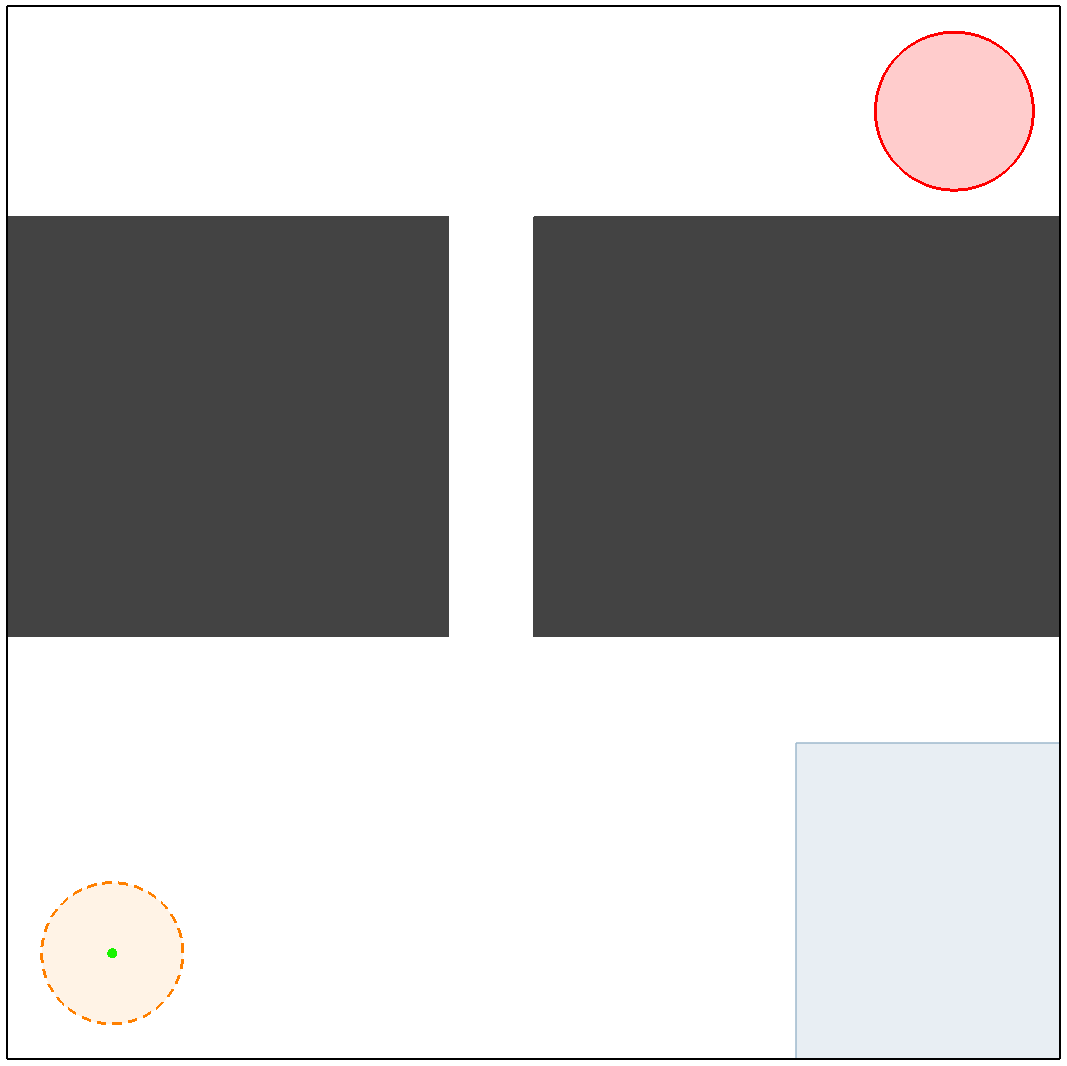}%
    \includegraphics[width=0.25\columnwidth]{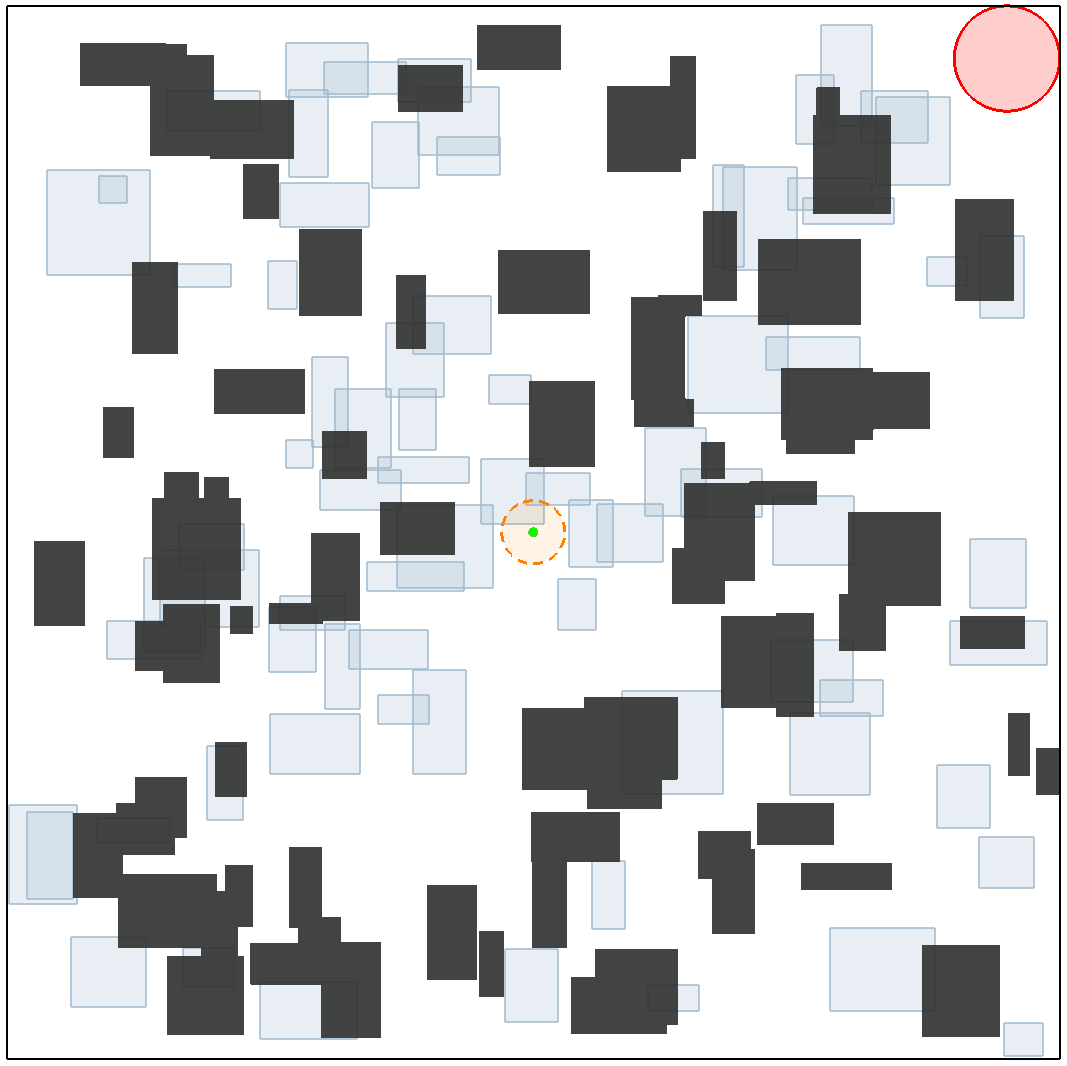}
    \caption{Simulated environments used for evaluation: from left to right, \emph{Empty}, \emph{Block}, \emph{Narrow}, and \emph{Random}. The start node and its covariance are denoted by a green dot and an orange circle, respectively, and the goal region by a red circle.}
    \label{fig:sim_maps_row}
    \vspace{-3mm}
\end{figure}

\section{Validation}
\subsection{Setup}
For each map from the previous section, we conducted 50 trials using pseudorandom seeds, with the same seed used across all five planners in each trial. Each planner was run for $100\,\mathrm{s}$ under the $\ell_2$ and $W_2$ objectives separately. The objective value was sampled every $1\,\mathrm{ms}$ by a separate monitoring thread, and the time to the first solution was recorded with sub-millisecond resolution.
All trials were executed sequentially with a memory limit of $4096\,\mathrm{MB}$ per run on an Intel Core i9-14900KF CPU running Ubuntu 22.04. The planners were implemented in C++ within OMPL~\cite{ompl2}. Our B-SST and B-RRT implementations were adapted from the source code accompanying~\cite{beliefA}, whereas RRBT was implemented following its original description \cite{RRBT}. Section~\ref{modifications} details the modifications and assumptions made in our adaptations.
\newcommand{\rl}[2]{%
  \multirow{#1}{*}{\rotatebox[origin=c]{90}{\textbf{#2}}}%
}
\begin{table*}[t]
\centering
\setlength{\tabcolsep}{3.0pt}
\setlength{\tabcolsep}{2.0pt}

\caption{
Planner performance across simulated maps and digital twins over 50 runs per map. $W_2$ and $\ell_2$ denote the optimization objectives. Metrics are mean first-solution cost/time, mean cost at 10/100\,$\mathrm{s}$, and success at 10/100\,$\mathrm{s}$. Means include finite solutions only; success is omitted for $\mathbb{R}^2$ maps, where all planners converge.
}

\label{tab:planner_benchmark}

\scriptsize
\begin{tabular*}{\textwidth}{
  |c|l|
  @{\extracolsep{\fill}}
  ccccc|
  ccccc|
  ccccc|
}

\cline{2-17}
\multicolumn{1}{c|}{}
& \multirow{2}{*}{\textbf{Planner}}
& Inf. BLT* & \MainAlgAccUninformed & B-RRT & B-SST & RRBT
& Inf. BLT* & \MainAlgAccUninformed & B-RRT & B-SST & RRBT
& Inf. BLT* & \MainAlgAccUninformed & B-RRT & B-SST & RRBT \\
\multicolumn{1}{c|}{}
&
& \textit{(Ours)} & \textit{(Ours)} & \cite{beliefA} & \cite{beliefA} & \cite{RRBT}
& \textit{(Ours)} & \textit{(Ours)} & \cite{beliefA} & \cite{beliefA} & \cite{RRBT}
& \textit{(Ours)} & \textit{(Ours)} & \cite{beliefA} & \cite{beliefA} & \cite{RRBT} \\
\cline{2-17}

\multicolumn{1}{c|}{}
& \textbf{Simulation $\mathbb{R}^2$}
& \multicolumn{5}{c|}{\textit{Empty}}
& \multicolumn{5}{c|}{\textit{Block}}
& \multicolumn{5}{c|}{\textit{Narrow}} \\
\cline{1-17}

\rl{4}{$W_2$}
& 1st time [$\mathrm{ms}$] & 0.14 & \textbf{0.12} & 4.11 & 2.51 & 3.72 & 8.77 & \textbf{8.35} & 12.37 & 26.93 & 14.09     & 19.30 & 19.19 & 9.72 & \textbf{6.15} & 249.23 \\
& 1st cost [$\mathrm{m}$]   & 140.45 & 140.45 & 186.36 & 185.03 & \textbf{140.40}     & 198.90 & 198.90 & 304.11 & 262.47 & \textbf{187.04}     & \textbf{212.50} & \textbf{212.50} & 325.11 & 292.49 & 215.40 \\
& 10\,$\mathrm{s}$ cost [$\mathrm{m}$] & \textbf{110.85} & 112.41 & 181.76 & 113.25 & 111.68    & \textbf{126.74}& 128.54 & 254.13 & 153.05 & 136.15      & 185.76 & 185.69 & 307.12 & \textbf{180.13} & 182.93 \\
& 100\,$\mathrm{s}$ cost  [$\mathrm{m}$]& \textbf{110.41} & 111.89 & 180.64 & 111.35 & 110.79    & \textbf{124.80} & 126.40 & 217.48 & 132.02 & 134.40     & 183.15 & 183.34 & 280.73 & \textbf{177.34} & 178.67 \\
\cline{1-17}

\rl{4}{$\ell_2$}
& 1st time  [$\mathrm{ms}$]& 0.14 & \textbf{0.12} & 4.05 & 2.44 & 4.18             & \textbf{9.04} & 9.06 & 12.31 & 24.96 & 14.49                      & 20.24 & 19.95 & 9.64 & \textbf{6.13} & 260.35 \\
& 1st cost  [$\mathrm{m}$] & \textbf{133.73} & \textbf{133.73} & 178.94 & 178.38   & 135.61 & 195.12 & 195.12 & 302.00 & 256.84 & \textbf{181.33}      & \textbf{208.18} & \textbf{208.18} & 321.96 & 288.25 & 212.26 \\
& 10\,$\mathrm{s}$ cost  [$\mathrm{m}$]& \textbf{106.10} & 106.21 & 174.49 & 108.14 & 107.21   & \textbf{122.84} & 123.99 & 251.23 & 135.33 & 132.80               & 181.15 & 181.11 & 303.95 & \textbf{176.50} & 179.70 \\
& 100\,$\mathrm{s}$ cost  [$\mathrm{m}$] & \textbf{105.97} & 106.08 & 173.39 & 106.65 & 106.34   & \textbf{120.99} & 121.84 & 214.31 & 124.18 & 131.34               & 178.51 & 178.74 & 277.54 & \textbf{173.93 }& 175.49 \\
\cline{1-17}

\multicolumn{1}{c|}{}
& \textbf{Digital Twin}
& \multicolumn{5}{c|}{\textit{Campus}}
& \multicolumn{5}{c|}{\textit{Office}}
& \multicolumn{5}{c|}{\textit{Rural}} \\
\cline{2-17}

\cline{1-17}
\rl{5}{$W_2$}
& 1st time  [$\mathrm{s}$] & \textbf{0.09} & \textbf{0.09} & 5.71 & 4.28 & 2.38    & \textbf{3.37} & 3.41 & -- & -- & 75.35                            & \textbf{0.01} & \textbf{0.01} & 0.03 & 0.05 & 0.02 \\
& 1st cost  [$\mathrm{m}$] & 186.90 & 186.90 & 183.21 & \textbf{168.93} & 180.35       & \textbf{506.36} & \textbf{506.36} & $\infty$ & $\infty$ & 506.86   & 165.52 & 165.51 & 212.97 & 222.47 & \textbf{158.92} \\
& 10\,$\mathrm{s}$ cost [$\mathrm{m}$] & 141.65 & \textbf{141.16} & 182.83 & 162.59 & 165.93       & 488.52 & \textbf{488.06} & $\infty$ & $\infty$ & $\infty$           & 132.74 & 140.92 & 206.83 & 160.53 & \textbf{127.66} \\
& 100\,$\mathrm{s}$ cost [$\mathrm{m}$] & \textbf{102.70} & 104.76 & 182.30 & 141.86 & 144.70       & \textbf{445.92} & 446.53 & $\infty$ & $\infty$ & 502.70            & 130.80 & 139.16 & 204.48 & 128.65 & \textbf{125.21} \\
& 10/100\,$\mathrm{s}$ success [\%] & 100/100 & 100/100 & 100/100 & 100/100 & 98/100       & 100/100 & 100/100 & 0/0 & 0/0 & 0/24                                  & 100/100 & 100/100 & 100/100 & 100/100 & 100/100 \\
\cline{1-17}

\rl{5}{$\ell_2$}
& 1st time  [$\mathrm{s}$] & \textbf{0.09} & \textbf{0.09} & 5.71 & 4.23 & 2.38       & 3.38 & \textbf{3.37} & -- & -- & 75.14                        & \textbf{0.01} & \textbf{0.01} & 0.03 & 0.05 & 0.02 \\
& 1st cost   [$\mathrm{m}$] & 181.16 & 181.16 & 168.25 & \textbf{156.39} & 172.50       & \textbf{496.09} & \textbf{496.09} & $\infty$ & $\infty$ & 492.49   & 162.51 & 162.51 & 210.00 & 219.41 & \textbf{156.22} \\
& 10\,$\mathrm{s}$ cost [$\mathrm{m}$]& \textbf{133.53} & 137.84 & 167.88 & 150.16 & 158.70       & 480.04 & \textbf{479.47} & $\infty$ & $\infty$ & $\infty$            & 128.87 & 136.34 & 204.01 & 158.15 & \textbf{125.01} \\
& 100\,$\mathrm{s}$ cost [$\mathrm{m}$]& \textbf{101.54} & 102.23 & 167.37 & 132.14 & 138.02       & 440.66 & \textbf{438.68}& $\infty$ & $\infty$ & 488.38            & 127.56 & 134.69 & 201.65 & 126.89 & \textbf{122.09} \\
& 10/100\,$\mathrm{s}$ success [\%] & 100/100 & 100/100 & 100/100 & 100/100 & 98/100       & 100/100 & 100/100 & 0/0 & 0/0 & 0/24                                  & 100/100 & 100/100 & 100/100 & 100/100 & 100/100 \\
\cline{1-17}
\end{tabular*}
\end{table*}
\newcommand{\rll}[2]{%
  \multirow{#1}{*}{\rotatebox[origin=c]{90}{\textbf{#2}}}%
}
\begin{table}[t]
\centering
\setlength{\tabcolsep}{6.0pt}
\setlength{\arrayrulewidth}{0.4pt}

\caption{
Planner performance on the \emph{Random} benchmark over 10 maps and 50 runs per map (500 runs per planner). $W_2$ and $\ell_2$ denote the optimization objectives. Metrics are mean first-solution cost/time and mean cost at 10/100\,$\mathrm{s}$.
}

\label{tab:random_benchmark}

\scriptsize
\begin{tabular}{
  |c|l|
  ccccc|
}

\cline{2-7}
\multicolumn{1}{c|}{}
& \multirow{2}{*}{\textbf{Planner}}
& Inf. BLT* & \MainAlgAccUninformed & B-RRT & B-SST & RRBT \\
\multicolumn{1}{c|}{}
&
& \textit{(Ours)} & \textit{(Ours)} & \cite{beliefA} & \cite{beliefA} & \cite{RRBT} \\

\cline{2-7}

\multicolumn{1}{c|}{}
& \textbf{Simulation $\mathbb{R}^2$}
& \multicolumn{5}{c|}{\textit{Random} $\times$ 10} \\
\cline{1-7}

\rll{4}{$W_2$}
& 1st time [$\mathrm{ms}$]                  & 3.30 & \textbf{3.27} & 6.56 & 4.98 & 45.32\\
& 1st cost [$\mathrm{m}$]                   & 99.97 & 99.97 & 124.89 &120.65 & \textbf{93.02} \\
& 10\,$\mathrm{s}$ cost [$\mathrm{m}$]      & 67.71 & 70.30 & 120.47 & \textbf{66.78} & 68.51 \\
& 100\,$\mathrm{s}$ cost [$\mathrm{m}$]     & 66.40 & 68.81 & 119.12 & \textbf{65.52} & 66.50 \\
\cline{1-7}

\rll{4}{$\ell_2$}
& 1st time [$\mathrm{ms}$]                  & 3.47 &	\textbf{3.30}	& 6.80 & 5.31 & 47.15 \\
& 1st cost   [$\mathrm{m}$]                 & 97.25  &	97.25	& 122.96	 & 119.81 & \textbf{90.28} \\
& 10\,$\mathrm{s}$ cost  [$\mathrm{m}$]     & 65.81  &	68.11	& 118.34 & \textbf{64.95} & 66.86 \\
& 100\,$\mathrm{s}$ cost [$\mathrm{m}$]     & 64.50	& 66.59	& 116.85 & \textbf{63.74} & 64.95 \\
\cline{1-7}
\end{tabular}
\end{table}
\subsection{Simulated Planning Environments}
We evaluated the five planners on 16 simulated maps: three custom maps in $\mathbb{R}^2$, ten random maps in $\mathbb{R}^2$, and three digital twins reconstructed from drone-based photogrammetry. Across 50 trials and two optimization objectives, this resulted in $16 \times 5 \times 50 \times 2 = 8000$ planner runs.

We modelled the robot as a point with observation-noise standard deviation $\sigma_{r}=0.1\,\mathrm{m}$ and bounded the sampled uncertainty by $\sigma_{\max}=5\,\mathrm{m}$.
The motion-model parameters were $u_{\max}=10\,\mathrm{m/s}$, $\sigma_{q}=\sqrt{0.001}\,\mathrm{m}$, and $\Delta\tau=0.1\,\mathrm{s}$.
These parameters limit the maximum displacement per propagation step to $1\,\mathrm{m}$, thereby reducing the number of intermediate belief updates while maintaining collision checks at intervals of at most $1\,\mathrm{m}$. We choose $P_{\text{inter}}$ such that $k_{\text{inter}}=3$, yielding a $3\sigma$ margin for collisions.

\subsubsection{Custom Maps in $\mathbb{R}^2$}
The three custom planar maps, inspired by~\cite{beliefA}, used for evaluation are shown in Fig.~\ref{fig:sim_maps_row}. \emph{Empty} admits a direct path to the goal and provides observations throughout the environment. \emph{Block} admits two solution classes, with the shorter path requiring the robot to gather information first. \emph{Narrow} requires prior information gathering to reduce the uncertainty to ensure traversal through the narrow corridor. We set $\sigma_{\mathrm{start}}=\sqrt{5}\,\mathrm{m}$ for all three maps.

\subsubsection{Random Maps in $\mathbb{R}^2$}
The \emph{Random} map in Fig.~\ref{fig:sim_maps_row} is representative of ten maps generated similarly to~\cite{bitstar2020IJRR}, each containing $75$ obstacles and $75$ measurement regions with dimensions sampled uniformly from $[2,10]\,\mathrm{m}$, and total coverage of each type limited to $1/3$ of the map area. The start state is centred in the $100\times100\,\mathrm{m}$ workspace with $\sigma_{\text{start}}=1\,\mathrm{m}$.

\subsubsection{Digital Twins}
We generated three digital twin maps shown in Fig.~\ref{fig:DTFIG}: \emph{Campus} is the smallest with dimensions $205 \times 124\,\mathrm{m}$; \emph{Rural} is slightly larger with dimensions $251 \times 138\,\mathrm{m}$ and sparse observable geometry consisting only of a few poles and one building; the \emph{Office} map is the largest at $438 \times 142\,\mathrm{m}$ and  consists of several buildings. 
Observable classes are buildings, fences, and poles for \emph{Campus} and \emph{Rural}, and buildings only for \emph{Office}.
Traversable classes are road, grass, ground, and dirt for \emph{Campus} and \emph{Rural}, and road only for \emph{Office}.
We simulate point cloud observations with a $20\,\mathrm{m}$ radius, centred around the robot at a height of $1.5\,\mathrm{m}$ above the ground. We set the robot radius to $0.25\,\mathrm{m}$.
The motion model parameters, observation noise, $\sigma_{\max}$, and $P_{\text{inter}}$ are the same as in the $\mathbb{R}^2$ maps.

\subsection{{Baseline Standardization}} \label{modifications}
Control-based BSP planners typically detect collisions using a predictive distribution with covariance $\mathbf{\Sigma}+\mathbf{\Lambda}$, where $\mathbf{\Lambda}$ captures uncertainty in the state estimate arising from future observations and closed-loop execution~\cite{RRBT,beliefA}. Because we consider open-loop execution without a feedback controller, we retain only the $xy$ marginal covariance $\mathbf{\Sigma}$, which generally produces less conservative plans. For a consistent comparison, all baselines use $\mathbf{\Sigma}$ for belief representation, propagation, and collision checking. We also augment them with point-cloud information matrices $\boldsymbol{\Omega}$, as described in Section~\ref{sec:icp_observation}, and restore isotropy after each update using the procedure in Section~\ref{sec:isotropicAssumption}. Control inputs are sampled by velocity magnitude, direction, and duration, with velocity bounded by $u_{\max}$.

\subsection{Results and Discussion}
The results in Table~\ref{tab:planner_benchmark}, Table~\ref{tab:random_benchmark}, and Fig.~\ref{fig:bench_dome} reveal two main trends. First, the proposed planners generally find solutions faster than the baseline methods. Second, informed sampling accelerates cost reduction when the latest solution defines a more restrictive informed subset. These effects are most apparent in the larger digital-twin environments.

The proposed planners achieved the fastest initial solution times on all maps except \emph{Narrow}, with particularly large improvements on \emph{Empty}, \emph{Campus}, and \emph{Office}. We attribute this primarily to the reachability formulation and information reuse discussed in Section~\ref{sec:Methodology}. In the digital twins, the baselines process observations at each time step along newly generated trajectories, making this advantage particularly pronounced. The gap narrows on \emph{Rural}, whose sparse point cloud, comprising a few poles and a small building, reduces this computational burden.

After an initial solution is obtained, the effect of informed sampling depends on how much of the original sampling domain can be excluded. In Fig.~\ref{fig:bench_dome}, the informed planner exhibits a faster reduction in median solution cost and discovers a shorter route through \emph{Campus} sooner. A similar improvement is observed on the random maps in Table~\ref{tab:random_benchmark}, where placement of the start and goal often produces an informed region occupying a smaller portion of the map.

The results also expose a trade-off between quickly obtaining a feasible solution and refining it within the available planning time. Although the proposed planners generally found their first solutions sooner, they did not consistently attain the lowest cost after $100\,\mathrm{s}$ on the random maps. \MainAlgAcc's mean final cost was, however, within $1.34\%$ of the lowest mean final cost obtained by any planner. 
In \emph{Narrow}, passage through the map requires substantial uncertainty reduction and appears more sensitive to the rewiring radius and the treatment of partial information during connection and rewiring.
\section{Conclusion \& Future Work}
Our planner provides fast initial solutions in simulated maps and more complex digital twins while achieving competitive final solution costs after 100$\,\mathrm{s}$. 

An important next step is to explore the optimality properties of our proposed method, including the tightness of the chosen informed region~\eqref{eq:informed_region_bound}. The empirical performance of Informed BLT* is comparable to, or better than, the near-asymptotically optimal B-SST planner \cite{beliefA}, suggesting that our adaptations retain favourable optimization behaviour in practice.

The digital twin experiments also demonstrate the scalability of the proposed approach: BLT* variants found solutions in all trials, whereas some baseline planners experienced substantial failures in the larger environments with more computationally expensive point-cloud observations.





\section*{ACKNOWLEDGMENT}
This work was supported by Defence Research and Development Canada (DRDC). ChatGPT (OpenAI) and Claude (Anthropic) were used for manuscript editing, code assistance for development, and figure generation.

\bibliographystyle{IEEEtran}
\bibliography{root}

\end{document}